\documentclass[11pt,a4paper]{article}
\usepackage[hyperref]{emnlp2018}
\usepackage{times}
\usepackage{amsmath,amsthm,amssymb}
\usepackage[vlined]{algorithm2e}
\usepackage{latexsym}
\usepackage[multiple]{footmisc}
\usepackage{graphicx}
 \usepackage{array}
\usepackage{tikz}
\usepackage{subfig}
\usepackage{url}
\usepackage{multirow}
\usepackage{pifont}% http://ctan.org/pkg/pifont

\newcommand{\ignore}[1] {}

\newcommand{\cmark}{\ding{51}\xspace}%
\newcommand{\xmark}{\ding{55}\xspace}%
\newcommand*\samethanks[1][\value{footnote}]{\footnotemark[#1]}

\aclfinalcopy

\title{WikiAtomicEdits: A Multilingual Corpus of Wikipedia Edits\\for Modeling Language and Discourse}

\author{Manaal Faruqui\Thanks{Both authors contributed equally.} \hspace{8mm} Ellie Pavlick\samethanks \hspace{8mm} Ian Tenney \hspace{8mm} Dipanjan Das\\
Google AI Language}

\begin{document}
\maketitle
\begin{abstract}

 We release a corpus of 43 million \textit{atomic edits} across 8 languages. These edits are mined from Wikipedia edit history and consist of instances in which a human editor has inserted a single contiguous phrase into, or deleted a single contiguous phrase from, an existing sentence.
 We use the collected data to show that the language generated during editing differs from the language that we observe in standard corpora, and that models trained on edits encode different aspects of semantics and discourse than models trained on raw, unstructured text. We release the full corpus as a resource to aid ongoing research in semantics, discourse, and representation learning.

\end{abstract}

\section{Introduction}

Written language often undergoes several rounds of revision as human authors determine exactly what information they want their words to convey. On Wikipedia, this process is carried out collectively by a large community at a rate of nearly two revisions per second \cite{yang-EtAl:2017:EMNLP2017}. While Wikipedia's revision history contains arbitrarily complex edits, our corpus and analysis focuses on \textit{atomic insertion edits}: instances in which an editor has inserted a single, contiguous span of text into an existing complete sentence (Table \ref{tab:data-examples}). This restriction allows us to make several assumptions which we believe make the data an especially powerful source of signal. Namely, we can assume that 1) some information was not communicated by the original sentence, 2) that information \textit{should} have been communicated (according to a human editor), and 3) that information \textit{is} communicated by the inserted phrase. Thus, we believe that a large data set of such edits is inherently valuable for researchers modeling inference and discourse and that the data can yield insights about representation at both the phrase and the sentence level.

\begin{table}[tb!]
  \centering
  \begin{tabular}{|p{.9\linewidth}|}
  \hline
  \textit{Adding new relevant information} \\
  She died there \textbf{in 1949} after a long illness.  \\\hline
\textit{Refining claim/Resolving ambiguity} \\  
%During a radio interview with HU2 Wrestling, 
Finlay announced he'd be on the 1000th episode of ``WWE Monday Night Raw''\textbf{, but he wasn't.} \\ \hline
\textit{Improving Discourse/Fluency}\\
It is \textbf{also} being evaluated as a potential biological control for the invasive plant $\dots$\\%garlic mustard.\\
%Mary's house, the northernmost, was erected first, begun in April of 1848 and completed \textbf{by} the following September. \\
  \hline 
  \end{tabular}
  \caption{Example atomic insertions (in bold) from the corpus and the types of semantic and discourse phenomena that such edits capture.}
  \label{tab:data-examples}
\end{table}

We mine Wikipedia edit history to create a corpus of 43 million atomic insertion and deletion edits covering 8 languages. We argue that the corpus contains distinct semantic signals not present in raw text. We thus focus our experiments on answering the following questions: 
 \begin{enumerate}
 \item How is language that is inserted during editing different from general Wikipedia text?
 \item What can we learn about language by observing the editing process that we cannot readily learn by observing only the final edited text? 
 \end{enumerate}
%\noindent
Specifically, the contributions of this paper are:
\begin{itemize}
\item A new corpus (WikiAtomicEdits) of 26M atomic insertions and 17M atomic deletions covering 8 languages (\S\ref{sec:editsdata} and \S\ref{sec:crowd}): {\small\url{http://goo.gl/language/wiki-atomic-edits}}.
\item Linguistic analysis showing that inserted language differs measurably from the language observed in general Wikipedia text (\S\ref{sec:corpus}). %E.g. we show that the edits capture explicitly information that might be communicated implicitly in general.
\item Language modeling experiments showing that models trained on WikiAtomicEdits encode different aspects of semantics and discourse than models trained on raw, unstructured text (\S\ref{sec:problem}). 
\end{itemize}

\section{Theoretical Motivation}
\label{sec:atomic-edits}

We borrow the idea of an \textit{atomic edit} from prior work in natural language inference, specifically natural logic \cite{lakoff,vanbenthem}. \newcite{maccartney} defines an atomic edit $e$ applied to a natural language expression $s$ as the insertion, deletion, or substitution of a sub-expression $p$ such that both the original expression $s$ and the resulting expression $e(s)$ are well-formed semantic constituents. E.g. $s=\textit{``She died from an illness''}$, $p=\textit{``in 1949''}$, and $e(s)=\textit{``She died in 1949 from an illness''}$. This formulation is desirable because it exposes a relationship between the surface form and the semantics of natural language while remaining agnostic about the underlying semantic representation. That is, the difference in ``meaning'' between $s$ and $e(s)$ is exactly the ``meaning'' of $p$ (in context), regardless of how that meaning is represented. 

We adopt this philosophy in creating our corpus. We focus our analysis specifically on atomic insertion edits. We make the assumption that editors on Wikipedia are attempting to communicate true information\footnote{This is true for the majority of edits, although about 13\% of edits are ``spam'' (\S\ref{sec:agreement}).} and to do so effectively. Insertion edits are thus particularly interesting because the underlying generation process admits the following assumptions:
%thus interesting in that they allow us to make the following assumptions:
\begin{enumerate}
\item The original sentence $s$ does not effectively communicate some piece of information.
\item A reasonable reader of $s$ would like/expect this information to be communicated.
\item This information is communicated by the inserted phrase $p$ (in the context of $e(s)$).
\end{enumerate}
%\noident
We therefore believe that the supervision provided by insertion edits can improve our understanding of semantics, discourse, and composition, and that the data released will be valuable for research in these areas. The goal of our experiments is to establish that the signal provided in these edits is distinct from what one could easily obtain given currently available text corpora. 

\section{WikiAtomicEdits: Corpus Creation}
\label{sec:editsdata}

\subsection{Extracting Edits}

Wikipedia edits can be accessed through Wikipedia dumps. The edits are stored as diffs on the entire Wikipedia page, meaning some processing is required to reconstruct the changes that were made at the sentence level. We use historical snapshots of each Wikipedia document and compare against subsequent snapshots to extract sentence-level edits. We strip the HTML tags and Wikipedia markup of the page and then run a sentence splitter \cite{gillick:2009} to obtain a list of sentences for each snapshot. Rather than run a full, quadratic-time \citep{myers1986ano} sequence alignment to compare the two lists of sentences, which is infeasible for long articles, we propose an efficient precision-oriented approximation. 

Given $n$ sentences in one snapshot (``base'') and $m$ sentences in a subsequent one (``edited''), we assume that most edits are local and restrict our attention to a fixed-size window. For each sentence $s_i$ in the base snapshot, %we assume $m \approx n \pm k$ and 
we compute pairwise BLEU scores \cite{papineni2002bleu} between $s_i$ and the sentences $\{t_j\}_{j=i-k}^{i+k}$ (\textbf{$k = 5$}) in the edited snapshot. We consider the sentence with the highest BLEU score in this window as a candidate. If the sentences are not identical and the difference consists of an insertion or deletion of a single contiguous phrase\footnote{We use the Python 2.7 \texttt{difflib} library to compute a minimal diff at the byte level.}, we add this example to the corpus. For each article, we run this algorithm over the most recent 100,000 snapshots as of February 2018. We extract edits for 8 languages. Statistics are shown in Table~\ref{tab:data-stats}.

\begin{table}[!tb]
  \centering
  \begin{tabular}{l|rrr}
  \hline
  Language & Ins & Del & Total \\\hline
  German & 3.3 & 1.9 & 5.2 \\
  English & 13.7 & 9.3 & 23.0 \\
  Spanish & 1.4 & 0.9 & 2.3 \\
  French & 2.0 & 2.0 & 4.0 \\
  Italian & 1.0 & 0.6 & 1.6 \\
  Japanese & 2.2 & 1.3 & 3.5 \\
  Russian & 1.4 & 0.9 & 2.3 \\
  Chinese & 0.7 & 0.4 & 1.1 \\
  \hline
  Total & 25.7 & 17.2 & 42.9 \\
  \hline 
  \end{tabular}
  \caption{The number of instances (in millions) of atomic insertions/deletions for each language.}
  \label{tab:data-stats}
\end{table}

\subsection{Insertions vs. Deletions}
\label{sec:ins-del}

We use the algorithm described above to extract both atomic insertions and atomic deletions. However, we chose to omit the deletions from our linguistic (\S\ref{sec:corpus}) and language modeling (\S\ref{sec:problem}) analyses for two reasons. First, our intuition is that spans which are deleted by an editor are more likely to be ``bad'' phrases (e.g. spam, false information, or grammatical errors introduced by a previous editor). To confirm this, we manually inspected 100 of each type of edit. We found that indeed deletions contained a higher proportion of spam text and malformed English (16/100) than did insertions (7/100).  Second, while insertions permit a clean set of assumptions about the relationship between the original sentence and the edited sentence (\S\ref{sec:atomic-edits}), it is more difficult to make generalizations about deletions. Specifically, it is difficult to say whether the original sentence \textit{should not} communicate the information in the deleted phrase (i.e. the phrase contains false, irrelevant, or otherwise erroneous information) or rather the original sentence/surrounding context \textit{already communicates} the information in the deleted phrase (i.e. the deleted phrase is redundant). As such, deletions are a noisier target for analysis. Nonetheless, we recognize that the deletions provide a related and likely useful signal. We thus include deletions in our corpus but leave their deeper linguistic analysis for future work.

\section{Corpus Quality \& Reproducibility}
\label{sec:crowd}

\subsection{Annotation}

Given the data collected as above, we now investigate whether the extracted edits are sufficiently clean to be useful for computational language analysis and modeling. To do this, we focus our attention specifically on the English, Spanish, and German subcorpora, as these are languages for which we could find a sufficient number of native speakers to perform the necessary annotation for our analysis. Thus, the discussion and results in this section may not be representative of the other languages in the corpus. 

We are interested specifically in two questions. First, we want to measure the overall corpus quality: how many of the inserted phrases represent meaningful edits and how many are simply noise (e.g. from editor or preprocessing error)? Second, we want to understand, at least in part, the reproducibility of the corpus: could we expect a different group of human editors to produce the same edits as those observed? 

To address these questions, we collect annotations in a semi-generative manner. Each annotator is shown a sentence $s$ and a phrase $p$ to be inserted, and is asked to insert $p$ into $s$ in order to form a new sentence $e(s)$. If $s$ is not a complete and well-formed sentence, or if there is no location at which $p$ can be inserted such that $e(s)$ would be a complete and well-formed sentence, annotators are instructed to mark the edit as an error. We use the ``error'' labels in order to study corpus quality (\S\ref{sec:qc}) and use the annotators' insertion location to estimate reproducibility (\S\ref{sec:agreement}).

We collect labels for 5,000 English edits, and 1,000 each for Spanish and German edits using a crowd-sourcing platform. We collect 5-way annotations for English and 3-way annotations for Spanish and German. Our choices of languages and the differing levels of redundancy were due to availability of annotators. We will release these 7,000 edits and their annotations with the corpus.%\ian{Can we say any more about the platform? I guess it counts as "proprietary"?}

\subsection{Corpus Quality}
\label{sec:qc}

To measure corpus quality, we compute the proportion of edits marked as errors by our annotators. Table \ref{tab:error-anno} shows our results. For English, in 78\% of cases our annotators agreed unanimously that $p$ could be inserted meaningfully into $s$ (55\% for Spanish; 85\% for German). These numbers reassure us that, while there is some noise, the majority of the corpus represents legitimate edits with meaningful signal. For more discussion of the errors refer to Supplementary Material.

\begin{table}[!tb]
    \centering
    \begin{tabular}{l|ccc}
    \hline
                        & en & es & de \\\hline
    No Error            & 78\% & 55\% & 85\% \\
    Possible Error      & 13\% & 30\% & 9\% \\
    Clear Error         & 9\% & 15\%  & 6\% \\
    \hline
    \end{tabular}
    \caption{Corpus quality for three languages for which we were able to collect annotations. ``No Error''/``Clear Error'' means annotators agreed unanimously that the edit was/was not an error; ``Possible Error'' means annotations were mixed.}
    \label{tab:error-anno}
\end{table}

\subsection{Agreement and Ambiguity}
\label{sec:agreement}

We next explore the extent to which the edits in the corpus are reproducible. In an ideal world, we would like to answer the question: given the same original sentences, would a different group of human editors produce the same edits? Answering this directly would require annotators with domain expertise and is infeasible in practice. However, we can use our crowdsourced annotation to answer a restricted variant of this question: given a sentence $s$ and an insertable phrase $p$, do humans agree on where $p$ belongs in $s$? We can measure agreement in this setting straightforwardly using exact match, and can interpret human performance as that of a ``perfect'' language model. I.e. we can interpret disagreement as evidence that reproducing the particular edit is dependent on exogenous information not available in the language of $s$ alone (e.g. knowledge of the underlying facts being discussed, or of the author's individual style).
%, and can view human agreement as an upper bound in an idealized scenario. That is

Based on our annotation experiment, we find that individual annotators agree with the original editor 66\% of the time for English, 72\% for Spanish, and 85\% for German.\footnote{We consider cases which the annotator marks as ``error'' to be a disagreement with the original editor.} More interesting than how often humans disagree on this task, however, is \textit{why} they disagree. To better understand this, we take a sample of 100 English sentences in which at least one human annotator disagreed with the original editor and no annotator marked the edit as an error. We then manually inspect the sample and record whether or not the annotators' choices of different insertion points give rise to sentences with different semantic meaning or simply to sentences with different discourse structure. 

In particular, we consider three categories for the observed disagreements: 1) the sentences are \textbf{meaning equivalent} from a truth-conditional perspective, 2) the sentences contain \textbf{significant differences in meaning} from a truth-conditional perspective, or 3) the sentences contain \textbf{minor differences or ambiguities} in meaning (but would likely be considered equivalent from the point of view of most readers). We also include an error category, for when the disagreement stems from a single annotator making an erroneous choice. Examples of each category are given in Table \ref{tab:disagreement-cats}. Note that the assessment of the truth conditions of the sentence and their equivalence is based on our judgment, and many of these judgments are subjective. We will release our annotations for this analysis with the corpus, to enable reproducibility and refinement in future research. 

\begin{table*}[ht!]
\centering
\begin{tabular}{p{.9\linewidth}}
% sentences lightly edited for presentation
\hline
\textit{\textbf{Meaning Equivalent}} \\
Paul Wheelahan\textbf{, the son of a mounted policeman,} was born in Bombala, South Wales$\dots$ \\
Paul Wheelahan was born in Bombala, South Wales\textbf{, the son of a mounted policeman,}$\dots$ \\\hline
%Paul Wheelahan\textbf{, the son of a mounted policeman,} was born in Bombala, New South Wales in 1930$\dots$ \\
%Paul Wheelahan was born in Bombala, New South Wales in 1930\textbf{, the son of a mounted policeman,}$\dots$ 
\textit{\textbf{Minor Difference / Ambiguity}}\\
She moved to Australia \textbf{in 1964} and attended the University of New South Wales$\dots$ \\
She moved to Australia and attended the University of New South Wales \textbf{in 1964}$\dots$ \\\hline
\textit{\textbf{Significant Difference in Meaning}} \\
$\dots$he and Bart have to share a raft with Ned Flanders and \textbf{his youngest son,} Todd Flanders. \\
$\dots$he and \textbf{his youngest son,} Bart have to share a raft with Ned Flanders and Todd Flanders. \\
%$\dots$he learns that he and Bart have to share the same raft with Ned Flanders and \textbf{his youngest son,} Todd Flanders. 
%$\dots$he learns that he and \textbf{his youngest son,} Bart have to share the same raft with Ned Flanders and Todd Flanders.
\hline
\end{tabular}
\caption{Examples of sentences falling into three disagreement categories, defined in terms of the truth conditions of the edited sentence. See text for a more detailed explanation.}
\label{tab:disagreement-cats}
\end{table*}

 Table \ref{tab:disagreement-counts} shows our results. We found 49\% to be meaning equivalent (i.e. the edit's location effected discourse structure only), and 22\% to have significant differences in meaning (i.e. the edit's location fundamentally changed the meaning of the sentence). An additional 13\% exhibited minor differences or ambiguities in meaning, and in the remaining 16\% of cases, the disagreement appeared to be due to annotator error. 

\begin{table}[ht!]
\centering
\begin{tabular}{l|r}
\hline
Meaning Equivalent & 49 \\
Significant Differences in Meaning & 22  \\
Minor Differences/Ambiguities & 13 \\
Annotator Error &  16  \\
\hline
\end{tabular}
\caption{Analysis of 100 sentences for which at least one annotator disagreed with the gold label and no annotator marked as an error.}
\label{tab:disagreement-counts}
\end{table}

\begin{table*}[ht!]
\centering
\begin{tabular}{lrl}
% lightly edited for presentation/compactness
\hline
Category  & Freq. & Example \\
\hline
Extend & 43\% & The population was 39,000 in 2004\textbf{, measured at 29,413 at the 2011 Census}. \\
%Extension &  Her father is of \textbf{Syrian and} Italian descent , and her mother is Danish and Sicilian. \\
Refine & 24\% & $\dots$began an investigation into Savile 's \textbf{apparent} history of abuse$\dots$\\
%Refinement & $\dots$a homologue of an algebra to be a model of the \textbf{equational} theory of that algebra$\dots$ \\ 
RE & 11\% & \textbf{Andrew} Sugerman has been involved in the production of motion pictures$\dots$\\
%Refer. Exp. & Mike Skinner won the only NASCAR Winston Cup \textbf{Series} exhibition race$\dots$\\
%Fluency & Its soldiers hoisted the Victory Banner atop \textbf{of} the building . \\
Fluency &  9\% & Philippine coconut jam\textbf{, meanwhile,} is made from coconut cream$\dots$\\
Error & 13\% & The team \textbf{are well - known as a loser team in the past 5 years.The team} is$\dots$ \\
%Error & His \textbf{is} first solo flight was after just two and one - half hours of demonstration.\\
\hline
\end{tabular}
\caption{High-level categories into which we manually characterize edits, to understand the variety of phenomena captured by the corpus. Frequencies are based on our annotation of a sample of 100 edits.}
\label{tab:edit-categories}
\end{table*}

\section{Corpus Linguistic Analysis}
\label{sec:corpus}

We now turn our attention to exploring the language in the corpus itself. In this section and in \S\ref{sec:problem}, our focus is on the questions put forth in the introduction: 1) how does the language that is inserted during editing differ from language that is observed in general? and 2) what can we learn about language by observing the editing process that we cannot readily learn by observing only raw text? Here, we explore these questions from a corpus linguistics perspective. The analysis in this section is based predominantly on the 14M insertion edits from the English subcorpus (Table \ref{tab:data-stats}). %\footnote{Similar trends are observed in other languages; see Supplementary Material.}  

\subsection{Manual Categorization of Insertions}
\label{sec:manual-classification}

We first characterize the types of insertions in terms of the function they serve. Manually inspecting the edits, we identify four high-level categories. Note that we do not intend these categories to be formal or exhaustive, but rather to be illustrative of the types of semantic and discourse phenomena in the corpus: i.e. to give sense of the balance between semantic, pragmatic, and grammatical edits in the corpus. The categories we identify are as follows: 

\begin{enumerate}
    \item \textbf{Extension}: the explicit addition of new information that the author of the original sentence did \textit{not} intend to communicate.\footnote{Whether or not the author ``intended'' to communicate something is based on our judgment. Since this annotation is intended to be exploratory, we allow a degree of informality.}
    \item \textbf{Refinement}: the addition of information that the author of the original sentence either intended to communicate or assumed the reader would already know. This category includes hedges, non-restrictive modifiers, and other clarifications or scoping-down of claims.
    \item \textbf{Fluency / Discourse}: grammatical fixes, as well as the insertion of discourse connectives (\textit{``thus''}), presuppositions (\textit{``also''}), and editorializations (\textit{``very''}).
    \item \textbf{Referring Expressions} (RE): changes in the name of an entity that do not change the underlying referent, such as adding a first name (\textit{``Andrew''}) or a title (\textit{``Dr.''}). RE edits could fall within our definition of ``refinement'', but because they are especially prevalent we annotate them as a separate category.
\end{enumerate}

\begin{figure*}[ht!]
    \centering
    \subfloat[English]{\includegraphics[width=.33\linewidth]{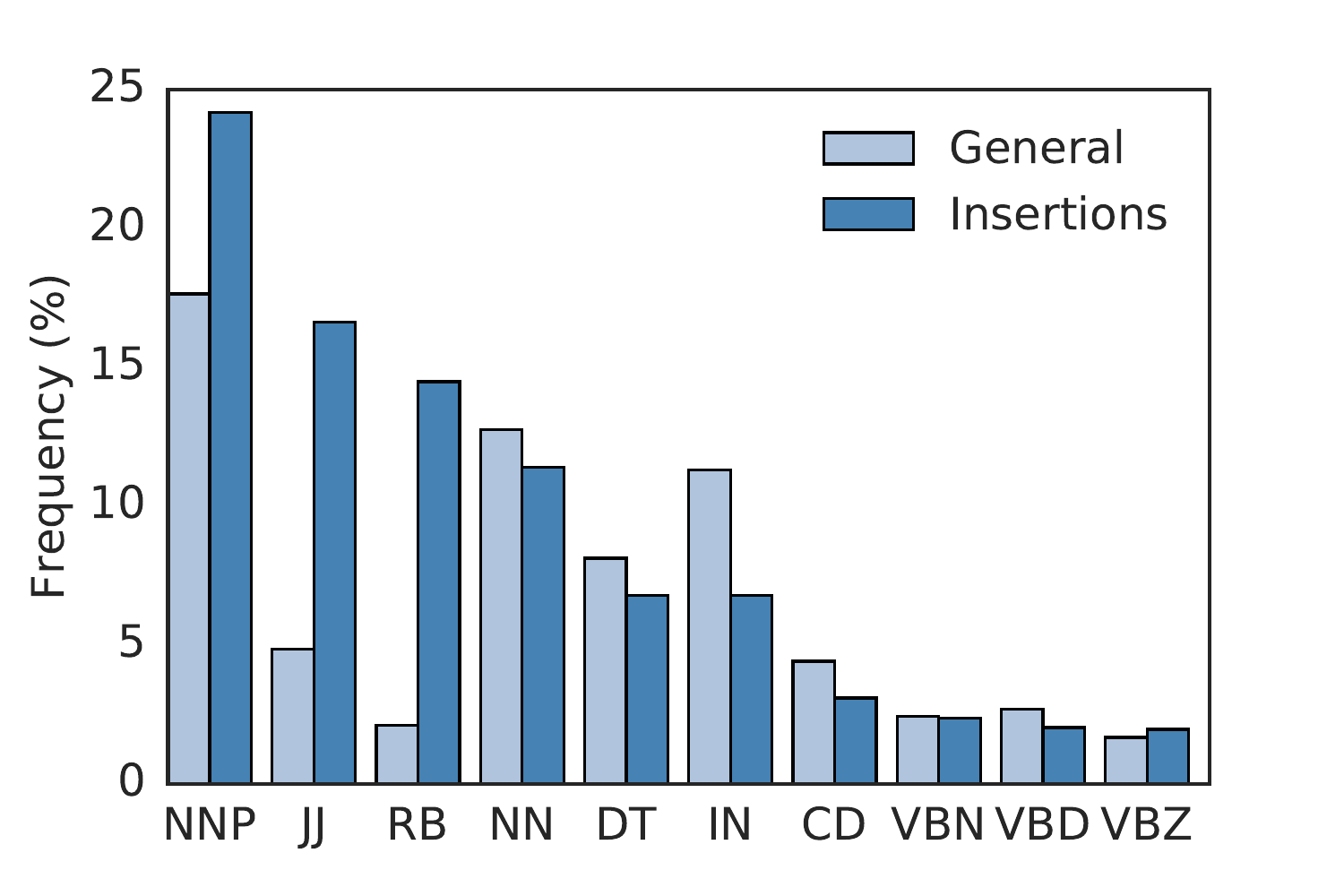}}
    \subfloat[Spanish]{\includegraphics[width=.33\linewidth]{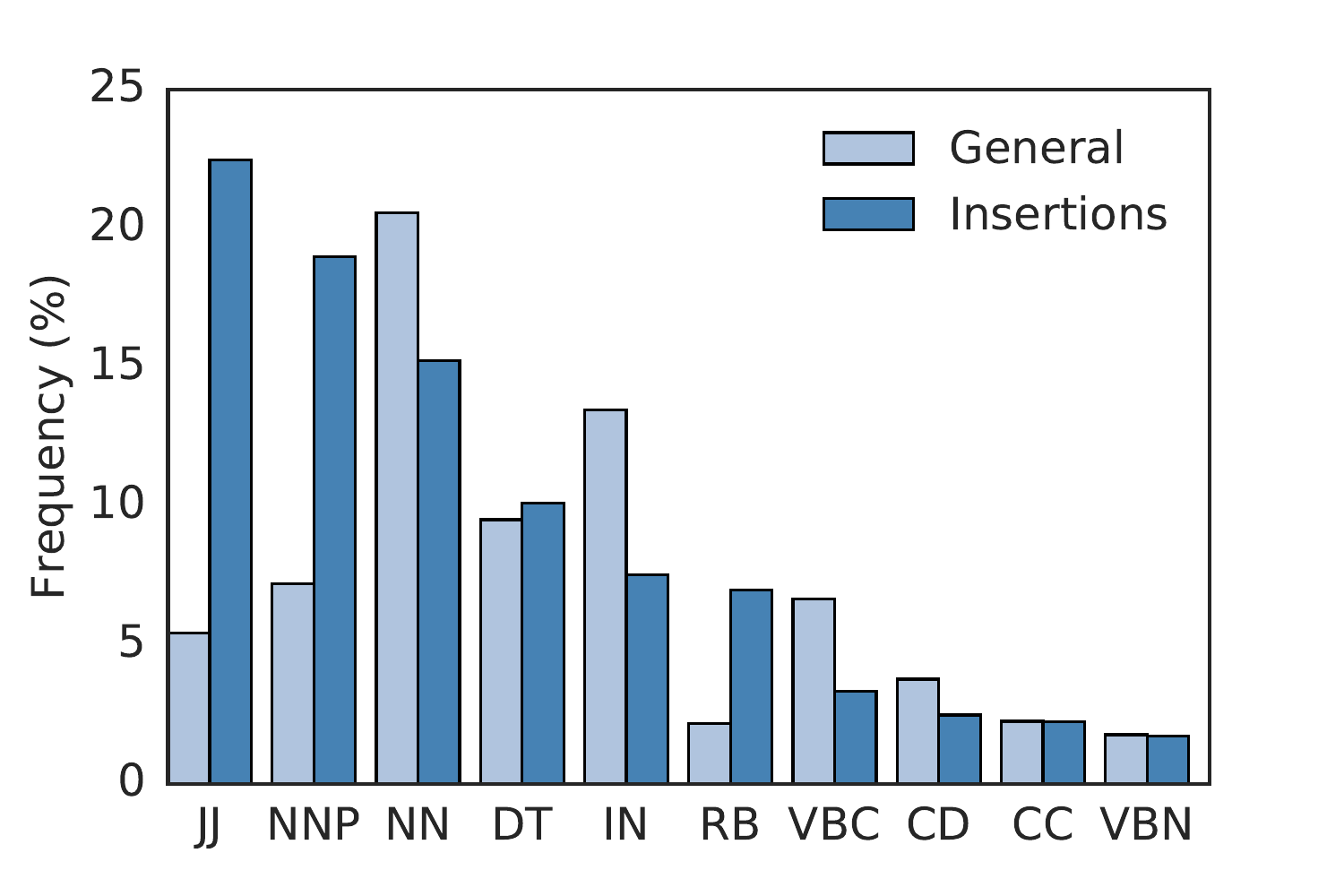}}
   \subfloat[German]{\includegraphics[width=.33\linewidth]{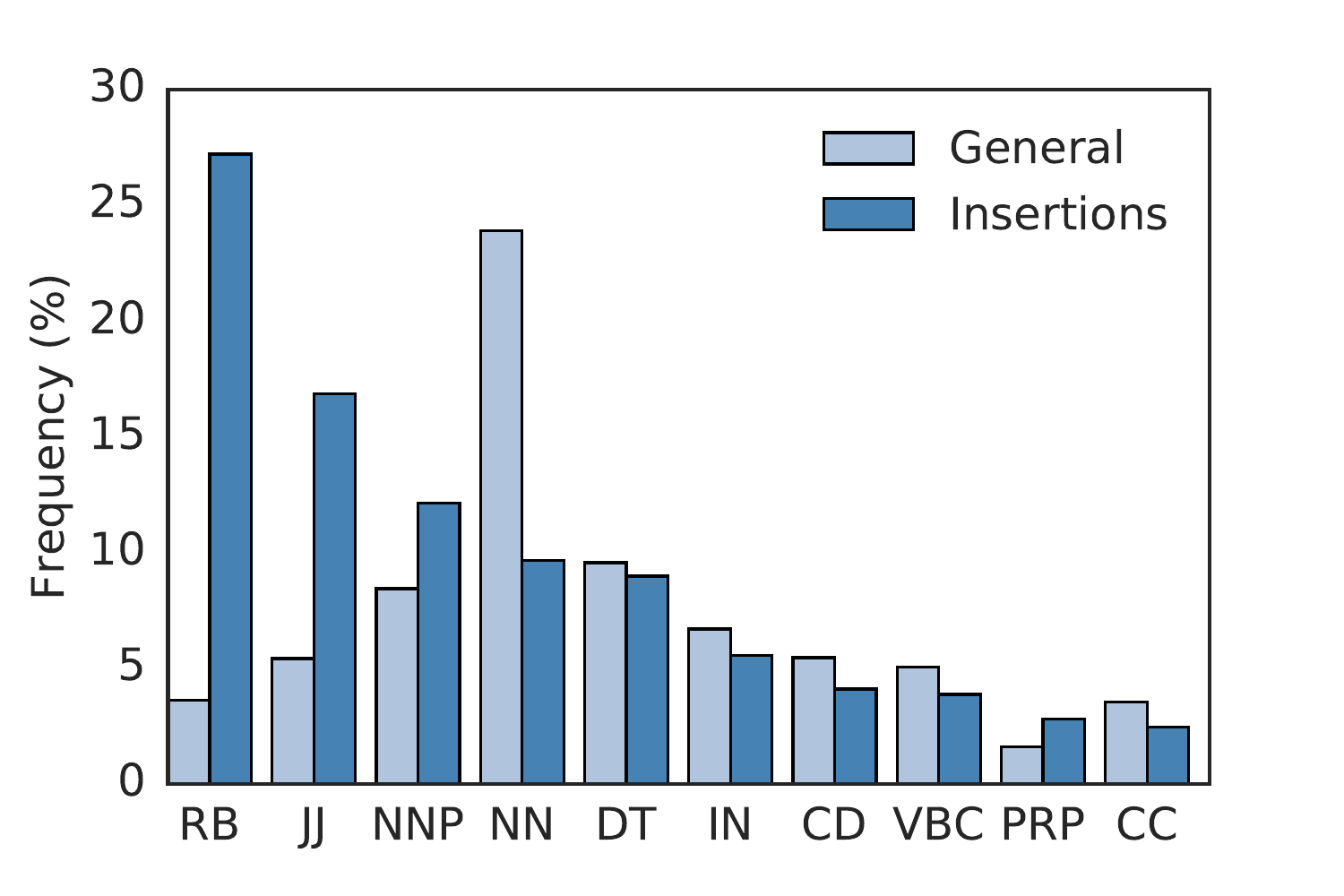}}
    \caption{Most frequent POS tags for English, Spanish, and German single-word insertions. Dark blue bars show the relative frequency among inserted phrases and light blue bars show the relative frequency among phrases observed in Wikipedia in general.}
    \label{fig:pos-dist}
\end{figure*}

\noindent We also define an \textbf{Error} category for spam, vandalism, and other ``mistake'' edits. 

We manually categorize 100 randomly-sampled edits. The breakdown is shown in Table \ref{tab:edit-categories}. In our sample, the majority (43\%) were extensions, and the second most frequent where refinements (24\%). No single category dominates and all are well-represented, suggesting that a variety of phenomena can be studied using this corpus.

\subsection{Comparing Insertions to Raw Text}

Understanding the high-level functions of edits, as above, provides some insight into the type of linguistic signals contained in the data. However, we are particularly interested in whether the language used for these functions is noticeably different from general Wikipedia text. That is: it is not obvious that the language humans use to e.g. extend or refine an existing claim should necessarily be different, in aggregate, from the language used to formulate these claims in general. We thus explore whether this is the case.

We first compare the distribution of parts of speech observed for the inserted phrases to the distribution of parts of speech that we observe in Wikipedia overall--i.e. in the sentences appearing in the final, published version of Wikipedia, not only the edit history. In order to compare the relative frequencies in a straightforward way, we look only at edits in which a single word was inserted.\footnote{In our corpus 30\% of inserted phrases are a single word, and 70\% are less than five words. We compared frequencies for longer POS sequences as well, but it did not yield particular insight over looking at single POS tags.} Figure \ref{fig:pos-dist} shows our results for English, Spanish, and German. We see, for example, that in English, adjectives and adverbs combined make up nearly 30\% of all inserted words, three and a half times higher than the frequency of adjectives/adverbs observed in the general Wikipedia corpus, and that proper nouns are inserted at a higher rate than would be suggested given their base frequency.

Looking more carefully, we see that the nature of the edits for each part of speech are qualitatively different as well. To explore this further, we look at which words appear at substantially higher rate as insertions than they do in the general Wikipedia corpus. We compute this as follows: for a word $w$ with part of speech $pos$, we compute the number of times $w$ occurs as an insertion per thousand insertions of any word of type $pos$, and compare this to the rate of occurrence of $w$ per thousand occurrences of any word of type $pos$ within the general Wikipedia corpus. Table \ref{tab:pos_rates} shows our results for English (Spanish and German are given in the Supplementary Material). In particular, we see that many words which are inserted at a significantly higher-than-baseline rate reflect ``refinement''-type edits. Many of these are words which the original author may have communicated implicitly but the editor chose to state explicitly, such as whether or not a person is a \textit{``current''}/\textit{``former''} public figure\footnote{We note that the addition of \textit{``former''} is likely tied to changes in the real world \cite{wijaya2015spousal}.} or is \textit{``famous''}. On the other hand, words which are inserted at a significantly lower-than-baseline rate are those which would be unlikely to be omitted by the original author. For example, if an event is famously the \textit{``first''} or the \textit{``only''} one of its kind, it is highly unlikely for the original author describing that event not to use these words explicitly. 

\begin{table}[!ht!]
    \centering
    \small
    \subfloat{
    \begin{tabular}{l|l|l|l}
    \hline
  &  \multicolumn{1}{c|}{NNP} & \multicolumn{1}{c}{JJ} & \multicolumn{1}{|c}{RB}\\\hline
 \parbox[t]{2mm}{\multirow{5}{*}{\rotatebox[origin=c]{90}{Over}}} &  City 16:2   & former 34:6    & also 187:91 \\
 &   Sir  7:1     & current 11:2  & currently 40:7 \\
 &   US 7:1      & famous 9:2        & very 24:11 \\
  &  John 6:3    & professional 10:3 & then 45:33 \\
   & Roman 4:1   & fictional 5:1     & allegedly 10:1 \\\hline
 \parbox[t]{2mm}{\multirow{5}{*}{\rotatebox[origin=c]{90}{Under}}} &    New 1:5     & first 9:29        & not 35:96 \\
 &   United 2:5  & only 2:20         & first 9:68 \\
  &  I 2:4     
9

  & other 12:26       & all  1:35 \\
   & de 2:4      & total 2:13        & only 22:47 \\
&    School 1:3  & such 3:13         & about 4:29 \\
    \hline
    \end{tabular}
    }\\
    \caption{Words that appear as insertions at significantly higher rates (top row) and significantly lower rates (bottom row) than their rate of occurance in Wikipedia in general. We compute ``rate'' as simply the observed occurrence of the given word per thousand occurrences of any word with the given POS. Table shows each word followed by (rate as insertion):(rate in general)} 
    \label{tab:pos_rates}
\end{table}

\section{Language Modeling Analysis}
\label{sec:problem}

We next explore the corpus from a language modeling perspective. Again, we are interested in understanding how the signal captured by the editing process is distinct from that captured by the final edited text alone, and in characterizing the types of signals we can learn from modeling the insertions directly. We investigate this through two tasks: first, given a sentence $s$ and insertable phrase $p$, predict the index $i$ at which $p$ should appear in $s$ (\S\ref{sec:model-location}), and second, given a sentence $s$ and an index $i$, generate candidate phrases that would be appropriate to insert into $s$ at $i$ (\S\ref{sec:model-phrase}).

\subsection{Predicting Insertion Locations}
\label{sec:model-location}

\paragraph{Task.} This task--given a phrase $p$ and a sentence $s$, choose the best index $i$ in $s$ at which to insert $p$--is identical to the task we asked humans to perform in \S\ref{sec:crowd}. We consider two simple models for performing this task: a basic language model and a discriminative model trained on the insertion data. We report performance as overall accuracy. We analyze whether a model which is trained to model insertions directly captures something different than a general language model in terms of the types of errors each model makes. 

\paragraph{Models.}

We evaluate two models. First, we evaluate a standard language modeling baseline (\textbf{General LM}), in which we simply insert the phrase $p$ at every possible point in $s$ and chose the index which yields the lowest perplexity. We use the LSTM language model from \newcite{jozefowicz2016exploring}, which obtained SOTA results on language modeling on the one billion words benchmark for English \cite{chelba2013one}. We train this  language model for each language on an average of $\sim$ 500 million tokens from Wikipedia. Second, we evaluate a discriminative model specifically trained on the insertion data (\textbf{Discriminative Model}). This model represents the base sentence using a sentence encoder that produces a context-dependent representation of every word index in the sentence, and then at test time, compares the learned representation of each index with the representation of the phrase $p$ to be inserted. We use a 256-dimensional 2-layer biLSTM encoder, initialized with FastText 300-dimensional word vectors \cite{mikolov2018advances,grave2018learning}.\footnote{\url{https://fasttext.cc/docs/en/crawl-vectors.html}} We hold out 50K and 10K insertion edits for each language as development and test sets, and use the remaining edits (insertions and deletions) as training data. This provides us with at least 1 million examples for training in each language (cf. Table~\ref{tab:data-stats}). See Supplementary Material for additional details.

\paragraph{Results.}
Table \ref{tab:insertionresults} shows the accuracy of each model for each language. We see that the discriminitve model trained on insertions directly performs better than the general LM by at least 1\% absolute accuracy on every language, and by 3.8\% absolute on average. It is worth emphasizing that this performance improvement is despite the fact that the general LM was trained with, on average, four times the number of tokens\footnote{The number of tokens in the WikiAtomicEdits is computed as the the total number of words in the edited sentence $e(s)$ after the insertion. Refer to Supplementary Material for more detailed statistics on the size of the dataset.} and is a much larger model--the general LM has $\sim$ 2 billion parameters \cite{jozefowicz2016exploring} compared to $\sim$ 1 million for the discriminative model.

\begin{table}[!tb]
  \centering
  \begin{tabular}{l|cc}
  \hline
        & General LM   & Discr. Model \\\hline
  German    & 68.1      & 72.9 \\
  English    & 58.7      & 68.4 \\
  Spanish    & 67.0      & 70.1  \\
  French    & 69.9      & 73.4  \\
  Italian    & 69.0      & 72.9 \\
  Japanese    & 73.0      & 74.2  \\
  Russian    & 72.9      & 74.3  \\
  Chinese    & 65.5      & 68.9  \\\hline
  Average   & 68.0      & 71.8 \\
  \hline
  \end{tabular}
  \caption{Insertion accuracy on the test set.} 
  \label{tab:insertionresults}
\end{table}

More interesting than raw performance is the difference in the types of errors that the models make. For each model, we take a random sample of 50 examples on which the model made a correct prediction and 50 examples on which the model made an incorrect prediction. We annotate these 200 examples\footnote{To avoid bias, the 200 examples are shuffled and the annotator does not know which group (correct/incorrect, or which model) each example belongs to.} according to the edit type classification discussed in \S\ref{sec:manual-classification}. Table \ref{tab:error-analysis} shows the results. We find a significant difference\footnote{We use the chi-squared test provided by scipy.stats.} ($p<0.01$) between the types of edits on which the General LM makes correct predictions and the types on which it makes incorrect predictions. Specifically, the General LM appears to be especially good at predicting location for fluency/discourse edits, and especially poor at predicting the location of refinement edits. In contrast, we do not see any significant bias in the errors made by the discriminative model compared to its correct predictions ($p=0.23$). We interpret this as evidence that the insertion data captures some semantic signal that is not readily gleaned from raw text corpora.

\begin{table}[!tb]
\centering
\begin{tabular}{l|c|cc|cc}
\hline
& Base & \multicolumn{2}{c|}{General} & \multicolumn{2}{c}{Discr.} \\
& Freq. & \multicolumn{2}{c|}{LM} & \multicolumn{2}{c}{Model} \\
         &   & \cmark & \xmark & \cmark     & \xmark \\\hline
Extend   & 25 & 21    & 19    & 25        & 21 \\ %43\%
Refine   &  14 & 7     & 18    & 13        & 14 \\ %24\%
RE        & 6 & 7     & 9     & 4         & 9 \\ %11\%
Fluency   & 5 & 15    & 4     & 8         & 6 \\ % 9\%
\hline
\end{tabular}
\caption{Relationship between model accuracy and insertion type, based on a sample of 50 correct (\cmark) and 50 incorrect (\xmark) predictions from each model. Base frequency is shown for reference and is based on our analysis from \S\ref{sec:manual-classification}. The General LM shows a bias in accuracy by insertion type. This bias is not observed for the discriminative model.}
\label{tab:error-analysis}
\end{table}
% chi statistics based scipy.stats.chisquare

\subsection{Predicting Insertion Phrases}
\label{sec:model-phrase}

\paragraph{Task.} In a final set of experiments, we explore a generative version of the language modeling task: given a sentence $s$ and an specified index $i$, generate a phrase $p$ which would be appropriate to insert into $s$ at $i$. We are interested in what such a model can learn about the nature of how sentences are extended: what type of information would be relevant from a semantic perspective, and natural from a discourse perspective to insert at a given point? We train two models for this task, one trained on the training split of the WikiAtomicEdits corpus, and one baseline trained on a comparable set of phrasal insertions not derived from human edits. We evaluate on the same 10K held-out insertion edits as in \S\ref{sec:model-location}, and measure performance using both a strict ``exact match'' as well as a softer similarity metric.

\paragraph{Model.}

We use an standard sequence-to-sequence model \cite{sutskever2014sequence}, modifying the input with a special token denoting the insertion point. For example, given the input [\textit{`` Angel '' is a song recorded by $<$ins$>$ pop music duo Eurythmics .}], the model would be trained to produce the target phrase [\textit{the British}]. We use a two-layer bidirectional encoder using the same 300-dimensional FastText embeddings as in \S\ref{sec:model-location}, and a sequence decoder with attention \cite{bahdanau+al-2014-nmt} using a learned wordpiece model \cite{schuster2012japanese} with a vocabulary of 16,000.

%We use an encoder-decoder with attention model as described in \newcite{bahdanau+al-2014-nmt}. We modify the model to take the index $i$ at which the generated phrase is to be inserted as an additional side input. We take the top-level encoder activations $e_{i}$ and $e_{i+1}$ from the left and right of the insertion point, run these through an additional single-layer encoder to produce a ``query'' encoding $q = biRNN([e_{i}, e_{i+1}])$ that directs the decoder to generate a phrase $p$ appropriate to that index. The model is described in greater detail in the Supplementary Material. %(\ref{sec:predicting-insertion-appendix}).
%\footnote{A simpler model would inject a special token into the source sentence, e.g. \texttt{"I like <ins> ice cream"} and use an ordinary seq2seq model. Our approach is more efficient in the case where a single sentence may have multiple insertion points.}

\begin{table*}[ht!]
    \centering
    \small
    \begin{tabular}{l|lll|l}
    \cline{1-2}\cline{4-5}
  \multicolumn{2}{c}{She is cited as the first female superstar of Hindi Cinema} && \multicolumn{2}{c}{He is married to Aida Leanca}\\
  \multicolumn{2}{c}{\textbf{and India 's Meryl Streep}} && \multicolumn{2}{c}{\textbf{and has two children}}\\
% \multicolumn{2}{c}{image saying ``Why did you insult the Pope?!''.} && \multicolumn{2}{c}{}\\
     \cline{1-2}\cline{4-5}

 \multicolumn{1}{c|}{Edits} & \multicolumn{1}{c}{General} &&  \multicolumn{1}{c|}{Edits} & \multicolumn{1}{c}{General} \\
     \cline{1-2}\cline{4-5}

   and is the best actress of the film & in japan   && and has a daughter & in january \\
 and is the best actress of the Indian cinema  & in june  &&, and has a daughter   & in june   \\
    and is the best actress of the film industry & in 2011 &&, and has a daughter and a daughter & in january 2012  \\
   %new & first &&, making it one of the $''$ $''$ $''$ $''$ singles albums & in 2010 \\
   %controversial & whole && , making it one of the top ten albums & of the UK  \\
    \cline{1-2}\cline{4-5}
    \end{tabular}
    \caption{Predicted phrase insertions from model trained on Edits vs. General corpus.  The Edits model better captures the discourse function of the human edit, e.g. elaborating on the previously-stated fact, while the General model gives syntactically-appropriate but generic insertions.}
    \label{tab:phrase-gen-examples}
\end{table*}

\begin{table}[ht!]
\centering
\begin{tabular}{l|cc}
\hline
                & Edits    & General \\\hline
Log Perplexity  & 8.32          & 9.23 \\
Exact Match     & 13.1\%        & 8.0\% \\
%MRR             & 2.41          & 2.68 \\ 
Similarity@1    & 0.54          & 0.48 \\
%Similarity@10   & 0.54          & 0.46 \\
%Similarity@Best & 0.69          & 0.61 \\
\hline
\end{tabular}
\caption{Comparison of how closely each model's generated phrases match the phrase inserted by the human editor. ``Edits'' was trained on WikiAtomicEdits and ``General'' was trained on comparable data not derived from human edits. We consider the top 10 phrases generated by each model.}
\label{tab:phrase-gen-results}
\end{table}

\paragraph{Experimental Design.}
We train one version of this model on the same set of 23M English examples as the discriminative insertion model from \S\ref{sec:model-location}; we refer to the model trained on this data as \textbf{Edits}. For comparison, we train an identical model on a set of simulated insertions which we create by sampling sentences from Wikipedia and removing contiguous spans of tokens, which we then treat as the insertion phrases. To ensure that this data is reasonably comparable to the WikiAtomicEdits data, we parse the sampled sentences \cite{andor2016globally} and only remove a span if it represents a full subtree of the dependency parse and is not the subject of the sentence.\footnote{Not all of the inserted phrases in WikiAtomicEdits are well-formed constituents. However, generating psuedo-edits using this heuristic provided a cleaner, more realistic comparison than using fully-random spans.} We generate 23M such ``psuedo-edits'' for training, the same size as the WikiAtomicEdits training set. We refer to the model trained on this data as \textbf{General}.

\paragraph{Results.}

We look at the top 10 phrases proposed by each model, as decoded by beam search. In addition to reporting standard LM perplexity, we compute two measures of performance, which are intended to provide an intuitive picture of how well each model captures the nature of the information that is introduced by the human editors. Specifically, we compute \textbf{Exact Match} as the proportion of sentences for which the model produced the gold phrase (i.e. the phrase inserted by the human editor) somewhere among the top 10 phrases. We also compute \textbf{Similarity@1} as the mean cosine similarity of each top-ranked phrase and respective gold phrase over the test set. We use the sum of the Glove embeddings \cite{glove} of each word in the phrase as a simple approximation of the phrase vector.

Table \ref{tab:phrase-gen-results} shows the results. We see that, compared to the model trained on General Wikipedia, the model trained on WikiAtomicEdits generates edits which are more similar to the human insertions, according to all of our metrics. Table \ref{tab:phrase-gen-examples} provides a few qualitative examples of how the phrases generated by the Edits model differ from those generated by the General model. Specifically, we see that the Edits model proposes phrases which better capture the discourse function of the human edit: e.g. providing context for/elaboration on a previously-stated fact. We note that this does not mean that training on Edits is inherently ``better'' than on General text, but rather that the supervision encoded by the WikiAtomicEdits corpus encodes aspects of language that are distinct from those easily learned from existing resources. %\ian{Re-wrote last sentence to be less hedgy, PTAL.}

\section{Related Work}
\label{sec:related-work}

\paragraph{Wikipedia Edits.} Wikipedia edit history has been used as a source of supervision for a variety of NLP tasks, including sentence compression and simplification \cite{Yamangil:2008,Yatskar:2010}, paraphrasing \cite{L10-1571}, entailment  \cite{zanzotto-pennacchiotti:2010:CCSR,Cabrio:2012}, and writing assistance \cite{Zesch:2012,cahill:2013,wiked2014}. User edits from Wikipedia and elsewhere have also been analyzed extensively for insight into the editing process and the types of edits made \cite{daxenberger:2012,daxenberger:2013,yang-EtAl:2017:EMNLP2017} and to better understand argumentation \cite{tan+lee:14}. Particular attention has been given to spam edits \cite{Adler:2011} and editor quality \cite{leskovec2010}. Our work differs in that WikiAtomicEdits is much larger than currently available corpora, both by number of languages and by size of individual languages. In addition, our focus on atomic edits should facilitate more controlled studies of semantics and discourse.

\paragraph{Sentence Representation and Generation.} We view the WikiAtomicEdits corpus as being especially valuable for ongoing work in sentence representation and generation, which requires models of what ``good'' sentences look like and how they are constructed. Recent work has attempted to model sentence generation by re-writing existing sentences, either using crowdsourced edit examples \cite{split-and-rephrase} or unsupervised heuristics \cite{guu2018edit}; in contrast, we provide a large corpus of natural, human-produced edits.

Also related is recent work in sentence representation learning from raw text \cite{kiros2015skip,peters2018deep}, bitext \cite{mccann2017learned}, and other supervised tasks including NLI \cite{conneau2017supervised}. Especially related is work on learning representations from weakly-labelled discourse relations \cite{nie2017dissent,jernite2017discourse}, as the WikiAtomicEdits corpus captures similar types of discourse signal.

\section*{Description of Data Release}

Our full corpus is available for download at {\small\url{http://goo.gl/language/wiki-atomic-edits}}. The data contains 26M atomic insertions and 17M atomic deletions covering 8 languages. All sentences (both the original sentence $s$, and the edited sentence $e(s)$) have been POS-tagged and dependency parsed \cite{andor2016globally} as well as scored using a SOTA LM \cite{jozefowicz2016exploring}. We also release the 5K 5-way human insertion annotations for English, and 1K 3-way annotations each for Spanish and German, as described in \S\ref{sec:crowd}. 

\section{Conclusion}

We have introduced the WikiAtomicEdits corpus, derived from Wikipedia's edit history, which contains 43M examples of atomic insertions and deletions in 8 languages. We have shown that the language in this corpus is meaningfully different from the language we observe in general, and that models trained on this corpus encode different aspects of semantics and discourse than models trained on raw text. These results suggest that the corpus will be valuable to ongoing research in semantics, discourse, and representation learning.

\bibliography{acl2018}
\bibliographystyle{acl_natbib}

\end{document}